\documentclass[a4paper, 11pt, conference]{ieeeconf} 

\IEEEoverridecommandlockouts

\usepackage{amssymb}
\usepackage{graphicx}
\usepackage{amsmath}
\usepackage{subfigure}
\usepackage[latin1]{inputenc}
\usepackage{color}
\usepackage{multirow}
\usepackage[hypcap=false]{caption}
\usepackage{rotating}
\usepackage{floatrow}

\usepackage{bm}
\usepackage{xspace}
\usepackage{amssymb}
\usepackage{amsmath}
\usepackage{amsfonts}
\usepackage{euscript}

\newcommand{\mN}{{\mathcal N}}

\def\QED{\mbox{\rule[0pt]{1.5ex}{1.5ex}}}

\def\registered{{\ooalign {\hfil\raise .05ex\hbox{\scriptsize
R}\hfil\crcr\mathhexbox20D}}}

\def\REgistered{{\ooalign
{\hfil\raise.09ex\hbox{\tiny \sf R}\hfil\crcr\mathhexbox20D}}}

\makeatletter
\DeclareRobustCommand\onedot{\futurelet\@let@token\@onedot}
\def\@onedot{\ifx\@let@token.\else.\null\fi\xspace}

\definecolor{gold}{rgb}{0.85,.66,0}

\graphicspath{{pics/}{figs/}{figs/segOverlays/}{artwork/}}

\usepackage{booktabs}
\title{Detect, Quantify, and Incorporate Dataset Bias: \\ A Neuroimaging Analysis on 12,207 Individuals}

\author{Christian Wachinger$^{1}$, Benjamin Gutierrez Becker$^{1}$, Anna Rieckmann$^{2}$\\ \\ 
$^1$Artificial Intelligence in Medical Imaging (AI-Med.de), KJP, LMU München, Germany \\
$^2$Department of Radiation Sciences, Ume{\aa} Univeristy, Sweden
}

\begin{document}
\maketitle
\thispagestyle{empty}
\pagestyle{empty}

\begin{abstract}
Neuroimaging datasets keep growing in size to address increasingly complex medical questions. 
However, even the largest datasets today alone are too small for training complex models or for finding genome wide associations. 
A solution is to grow the sample size by merging data across several datasets. 
However, bias in datasets complicates this approach and includes additional sources of variation in the data instead. 
In this work, we combine 15 large neuroimaging datasets to study bias. 
First, we \emph{detect} bias by demonstrating that scans can be correctly assigned to a dataset with 73.3\% accuracy. 
Next, we introduce metrics to \emph{quantify} the compatibility across datasets and to create embeddings of neuroimaging sites.  
Finally, we \emph{incorporate} the presence of bias for the selection of a training set for predicting autism. 
For the quantification of the dataset bias, we introduce two metrics: the Bhattacharyya distance between datasets and the age prediction error. 
The presented embedding of neuroimaging sites provides an interesting new visualization about the similarity of different sites. 
This could be used to guide the merging of data sources, while limiting the introduction of unwanted variation. 
Finally, we demonstrate a clear performance increase when incorporating dataset bias for training set selection in autism prediction. 
Overall, we believe that the growing amount of neuroimaging data necessitates to incorporate data-driven methods for quantifying dataset bias in future analyses. 
\end{abstract}

\section{Introduction}
As neuroimaging is joining the ranks of a "big data" science with more and larger datasets becoming available~\cite{smith2018statistical}, the issue of dataset bias is becoming prevalent. 
In general, bias refers to statistics that are systematically different from the population parameters. 
In a collection of unbiased datasets, similar results should be achieved by running independent analyses on each dataset and it would be straightforward to pool subjects across datasets without introducing unwanted variation. 
Further, models learned on one dataset would naturally generalize to other datasets. 
However, in practice, neuroimaging datasets are subject to various types of biases. 
These include subject selection, acquisition method, and processing biases.
While efforts have been made and are still ongoing to improve image processing to limit the impact of dataset bias in the outcome (e.g., volume or thickness measurements), substantial bias still remains~\cite{guadalupe2017human,jovicich2009mri,kruggel2010impact,lewinn2017sample,nugent2013automated}. 


\emph{Selection bias} stems from the fact that subjects included in the study do not represent the overall population. 
Examples are (i) the recruitment of particular target groups, e.g., young adults; (ii) the recruitment of a particular disease group; or (iii) an over-representation of more educated participants in convenience samples. 
While the first two are potentially related to the study objective and can be controlled for, the third one is more difficult to control and also seems to appear in epidemiological studies~\cite{smith2018statistical}. 
A second bias stems from the \emph{image acquisition}, where magnetic field strength, manufacturer, gradients, pulse sequences and head positioning cause variations in the images.
While standardization efforts are undertaken for instance by the ADNI~\cite{jack2008alzheimer}, variation related to the scanner remains~\cite{kruggel2010impact}, and it is questionable if a further standardization is in the manufacturer's interest. 
Finally, there is \emph{processing bias} in image segmentation and registration, which is in part related to acquisition bias.  
The type of segmentation method and the selected parameters can largely influence the outcome. 
Further, head motion, voxel size and image noise can cause bias in segmentation results.  

In this paper, we first detect the magnitude of dataset bias present in large neuroimaging studies.
Instead of trying to remove the bias, we propose to incorporate it in the analysis, which requires to quantify it first.  
To this end, we introduce two dataset metrics: the Bhattacharyya distance in feature space and the age prediction error for quantifying model generalization by including a variable from subject demographics. 
In addition to operating on the level of datasets, we also look at a more fine-grained analysis on acquisition sites. 
Based on the dataset metric, we create an embedding of neuroimaging sites to visualize the similarity among them. 
Finally, we demonstrate the benefit of composing a training set based on the dataset metric for autism prediction.  


%
%

\section{Data}

We work on MRI T1 brain scans from 15 large-scale public datasets: ABIDE I+II~\cite{ABIDE}, ADHD200~\cite{ADHD200}, ADNI~\cite{jack2008alzheimer}, AIBL~\cite{AIBL}, COBRE~\cite{COBRE}, CORR~\cite{zuo2014open}, GSP~\cite{GSP}, HBN~\cite{alexander2017open}, HCP~\cite{HCP}, IXI\footnote{http://brain-development.org/ixi-dataset/}, MCIC~\cite{MCIC}, NKI~\cite{nooner2012nki}, OASIS~\cite{OASIS}, and PPMI~\cite{PPMI}. 
All datasets were processed with FreeSurfer~\cite{fischl2002whole} version 5.3. 
We keep only one scan per subject from longitudinal or test-retest datasets. 
After exclusion of scans with processing errors and incomplete meta data, we work with scans from 12,207 subjects (6,827 male; 8,126 controls; mean age: 33.5 (sd=23.9)).
Demographics per dataset are reported in Table~\ref{tab:dataStats}. 

\begin{table*}[t]
\center
  \begin{tabular}{lrrrrrrr}
  \toprule
Dataset	&	Diagnosis	&	  $N$	& \ \ \ Age (mean)  	& \ \ 	Age (SD)	& \ \ 	Males \%	 & \ \ 	Sites	& \ \	Patients	\\
\midrule															
ABIDE I	&	Autism	&	\ \ \ \ \ 1,095	&	17.1	&	8.1	&	85.2	&	24	&	526	\\
ABIDE II	&	Autism	&	1,032	&	15.2	&	9.4	&	76.1	&	17	&	477	\\
ADHD200	&	ADHD	&	965	&	12.1	&	3.3	&	61.8	&	8	&	384	\\
ADNI	&	Alzheimer's	&	1,682	&	73.6	&	7.2	&	55.0	&	62	&	1144	\\
AIBL	&	Alzheimer's	&	262	&	72.9	&	7.6	&	47.3	&	2	&	91	\\
COBRE	&	\ \ \ Schizophrenia	&	146	&	37.0	&	12.8	&	74.7	&	1	&	72	\\
CORR	&		&	1,476	&	25.9	&	15.4	&	50.0	&	32	&	0	\\
GSP	&		&	1,563	&	21.5	&	2.8	&	42.3	&	5	&	0	\\
HBN	&	Psychiatric	&	689	&	10.7	&	3.6	&	59.7	&	2	&	497	\\
HCP	&		&	1,113	&	28.8	&	3.7	&	45.6	&	1	&	0	\\
IXI	&		&	561	&	48.6	&	16.5	&	44.6	&	3	&	0	\\
MCIC	&	Schizophrenia	&	194	&	33.1	&	11.5	&	71.6	&	3	&	104	\\
NKI	&	Psychiatric	&	624	&	38.4	&	22.5	&	39.1	&	1	&	268	\\
OASIS	&	Alzheimer's	&	415	&	52.8	&	25.1	&	38.6	&	1	&	100	\\
PPMI	&	Parkinson's	&	390	&	61.2	&	10.0	&	62.6	&		&	284	\\
  \bottomrule
 \end{tabular}
 \caption{Demographics of 15 neuroimaging datasets.} \label{tab:dataStats} 
\end{table*}

\section{Name That Dataset} 

\begin{figure*}[t]
\begin{center}
	\includegraphics[width=0.49\textwidth]{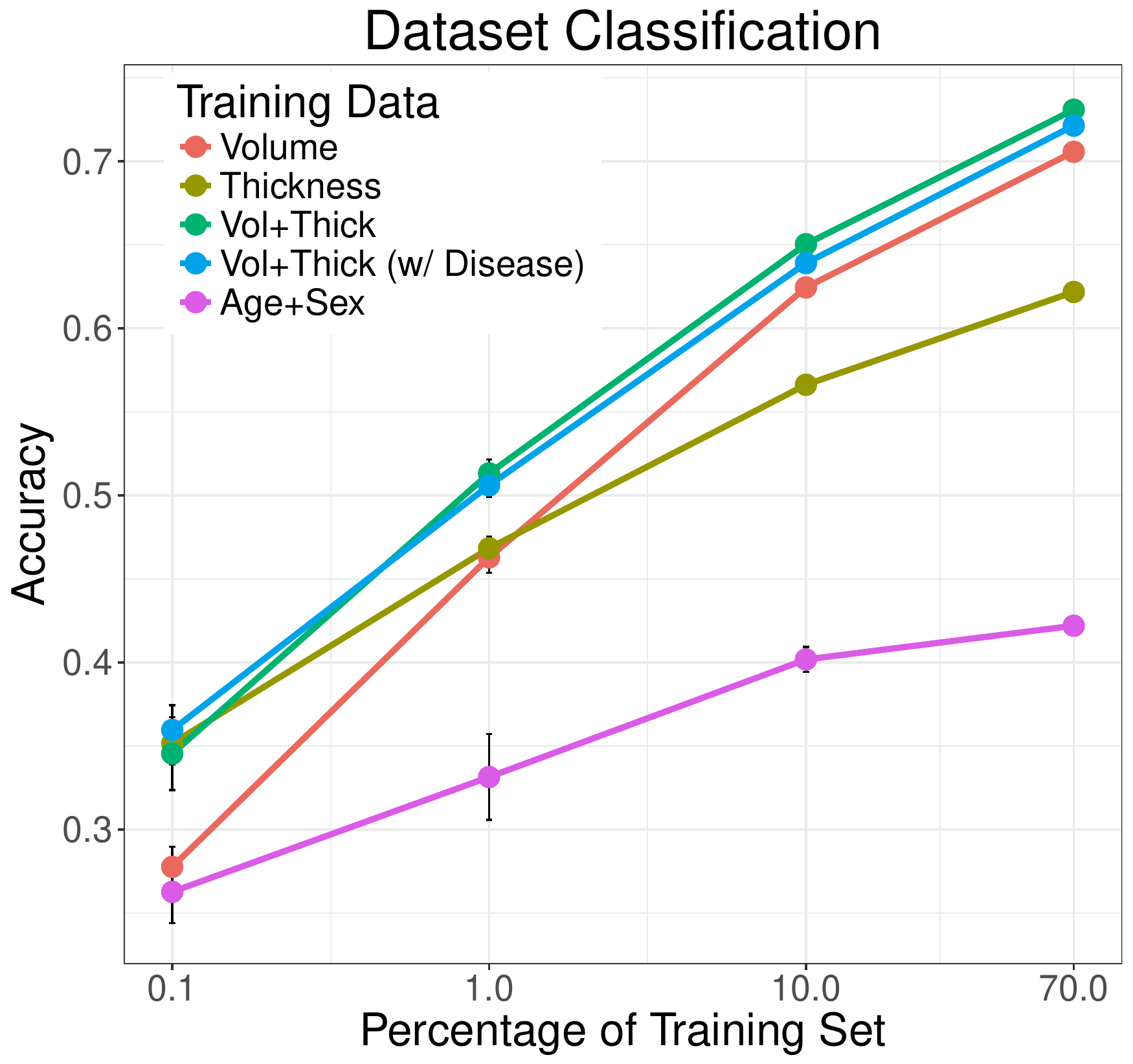} 
	\includegraphics[width=0.5\textwidth]{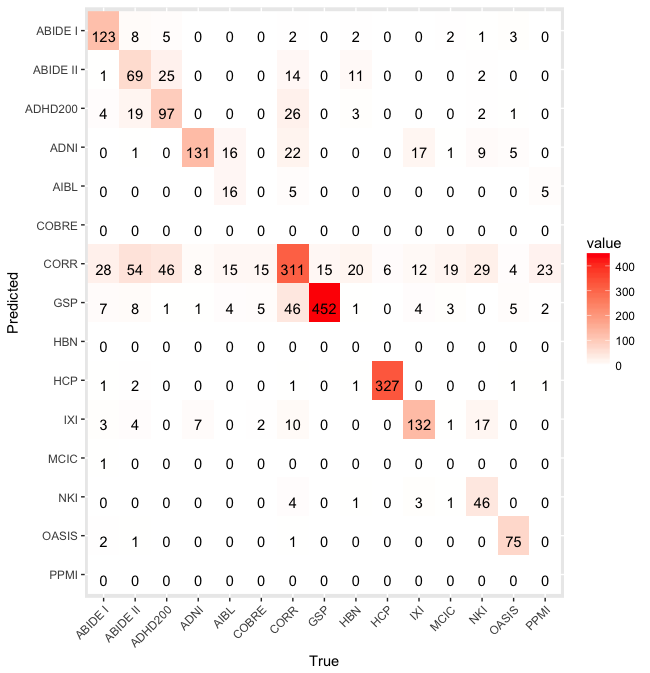} 
\caption{Left: Dataset classification accuracy for volume, thickness, and the combination of both, together with age and sex. The percentage of the data used for training is shown in log scale. Curves show the average score over 50 repetitions, error bars show the standard deviation. Right: Confusion matrix for volume and thickness with 70\% training data. 
\label{fig:DatasetClass}
}
\end{center}
\end{figure*}

In order to evaluate the impact of dataset bias, we  play the game \emph{Name That Dataset} on neuroimaing data that was originally proposed by Torralba and Efros~\cite{torralba2011unbiased} on natural images. 
The task is to predict the dataset a scan is coming from solely based on image measurements. 
Fig.~\ref{fig:DatasetClass} illustrates the performance for classifying the 15 datasets for different image features. 
A random forest classifier with default settings was used for the prediction~\cite{breiman2001random}. 
The splitting of training and testing dataset is done under consideration of the dataset. 
The performance of image-based classifiers increases logarithmically with the amount of training data. 
If no dataset bias was present, the prediction accuracy should be close to random chance (6.7\% for 15 datasets).
As not all datasets have the same size and have different distributions of age and sex, we compare to results of a classifier trained on age and sex as baseline. 
With only 0.1\% of the data used for training, volume measures perform similar to prediction with meta data. 
As we increase the amount of training data to 70\%, the accuracy increases over 73.3\% for the combination of volume and thickness features, which perform better than each of them alone. 
Compared to 42.2\% for age and sex, this illustrates that there is a strong bias in the datasets that cannot be explained by basic demographics.  
We focused the analysis on selecting only healthy controls, because we thought that the inclusion of patients would facilitate the classification. 
However, the results are similar, as shown for the combination of volume and thickness in Fig.~\ref{fig:DatasetClass}. 



From the confusion matrix, we see that datasets with a similar population result in higher confusion, e.g., between ABIDE I, ABIDE II, and ADHD200. 
Single site datasets like HCP are very homogeneous and do therefore show almost no confusion with any of the other datasets. 
In contrast, multi-site datasets like CORR that also cover a wide age range, show  high confusion with other datasets. 
Overall, however, high classification accuracy and the strong diagonal indicate that datasets possess unique, identifiable characteristics. 

The lesson learned from this experiment is that even when working with image-derived values that represent physical measures (volume, thickness), there is still substantial bias in datasets, although techniques like atlas renormalization~\cite{han2007atlas} were employed to improve consistency across scanners. 
Of course, much of the bias can be attributed to the different goals of the studies, like the inclusion of subjects from different age groups. 
However, even when focusing on datasets that cover a similar age range, we observe a high accuracy. 
While we are not aware of previous attempts on trying to \emph{Name That Dataset}, our results echo concerns raised in previous studies. 
In a large ENIGMA study of over 15,000 subjects on brain asymmetry \cite{guadalupe2017human}, it was reported that dataset heterogeneity explained over 10\% of the total observed variance per structure. 
On the ADNI, with an optimized MPRAGE imaging protocol across all sites~\cite{jack2008alzheimer}, the intra-subject variability of compartment volumes for scans on different scanners was roughly 10 times higher than repeated scans on the same scanner~\cite{kruggel2010impact}.  
Similarly, previous studies reported on a drop of accuracy when training on different datasets~\cite{wachinger2016domain} or working with multi-site data~\cite{nielsen2013multisite}.


\section{Quantifying Dataset Compatibility}

\subsection{Compatibility Metrics}
Having shown the presence of dataset bias, our next aim is to define metrics that quantify their compatibility. 
Given data sources $A$ and $B$,  the metric $m(A,B)$ expresses the compatibility among them. 
As first metric, we propose to compute the Bhattacharyya distance between data sources. 
To this end, we estimate multivariate normal distributions $\mN_A$ and $\mN_B$, respectively, and compute the Bhattacharyya distance between them $m(A,B) = d_B(\mN_A,\mN_B)$. 
The dimensionality of the distributions corresponds to the number of image-derived measures, where we use brain structure volumes in our experiments. 

As second metric, we propose to compute the age prediction error, which includes a variable from subject demographics (age). 
We train an age regression model on the source set $A$ and predict on the target set $B$. 
Since we know the chronological age on the target set, we compute the average mean age prediction error, $\varepsilon(A,B)$.
To have a symmetric metric, we set $m(A,B) = \frac{1}{2} ( \varepsilon(A,B) + \varepsilon(B,A) )$. 
Age estimation has previously been used for modeling healthy aging and differentiating it to abnormal aging in dementia~\cite{franke2010estimating,becker2018gaussian} and has the advantage, in contrast to other prediction tasks, that age is a commonly available variable. 
We use random forest regression on volume measures  for the age regression. 
While the Bhattacharyya distance is measuring the similarity of image features, the age prediction error expresses how well one data source is suited for training a model that is deployed on a second dataset. 

\begin{figure*}[t]
\begin{center}
	\includegraphics[width=1.1\textwidth]{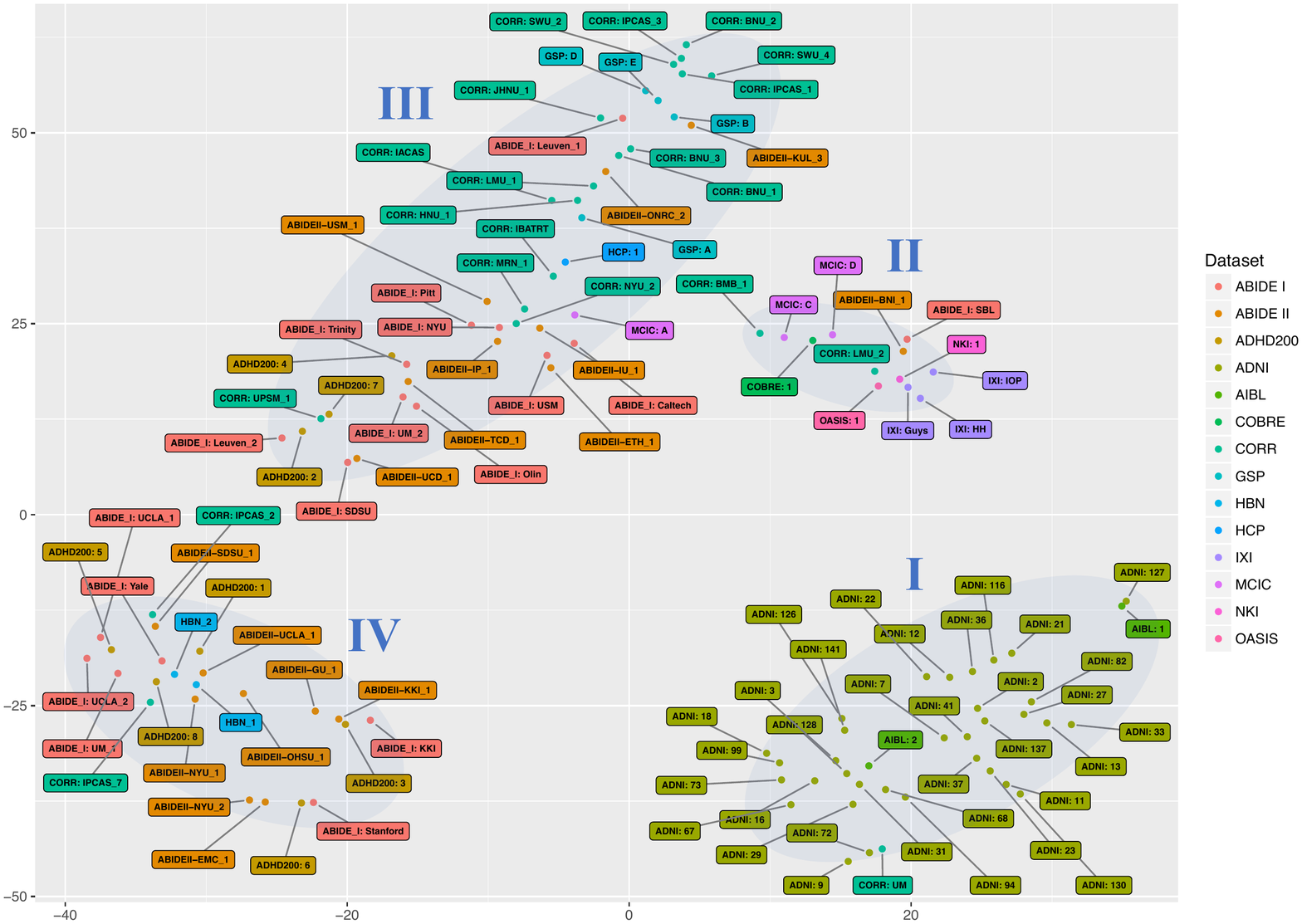} 
\caption{Embedding of neuroimaging sites. Color encodes dataset. Name of site and dataset displayed next to point. Four clusters are highlighted. 
\label{fig:Embedding}
}
\end{center}
\end{figure*}

\subsection{Site Embedding}
To investigate the similarity across datasets, we create an embedding based on the metrics. 
However, many of the large neuroimaging datasets are \emph{multi-site datasets}, i.e., scans were acquired at different scanning sites. 
Some initiatives like the ADNI put major efforts in the standardization of scans across sites. 
Other multi-site datasets like ABIDE~\cite{ABIDE} retrospectively aggregate data that was independently acquired from laboratories around the world. 
To study the variation in such datasets, we perform an analysis of variance (ANOVA) on the ABIDE I dataset with age, age squared, sex, diagnosis, and site as variables. 
For putamen, amygdala, and nucleus accumbens, the percentage of variance explained by site is 20.9\%, 23.7\%, and 32.7\%, respectively, while the total variance explained from all variables ranged between 32.9\% to 38.7\%. 
Site is therefore the major source of variation, several times higher than age, sex, or diagnosis. 
Based on these results, we will operate on the level of sites, instead of datasets, in the following. 

We compute the metric $m$ between all pairs of sites in our data, where we limit the analysis to sites with more than 25 subjects to have enough samples for a reliable estimation.
Based on the pair-wise age prediction across all sites, we use the resulting distance matrix in t-SNE~\cite{maaten2008visualizing} for visualizing the similarity of sites. 
Fig.~\ref{fig:Embedding} shows the embedding, where the age prediction error was used as metric and the perplexity in t-SNE was set to 5. 
We only show results for the age prediction error in this experiment because it yielded a clearer separation of datasets. 
We compare both metrics in section~\ref{sec:autism}.

\begin{figure*}[t]
\begin{center}
	\includegraphics[width=1.1\textwidth]{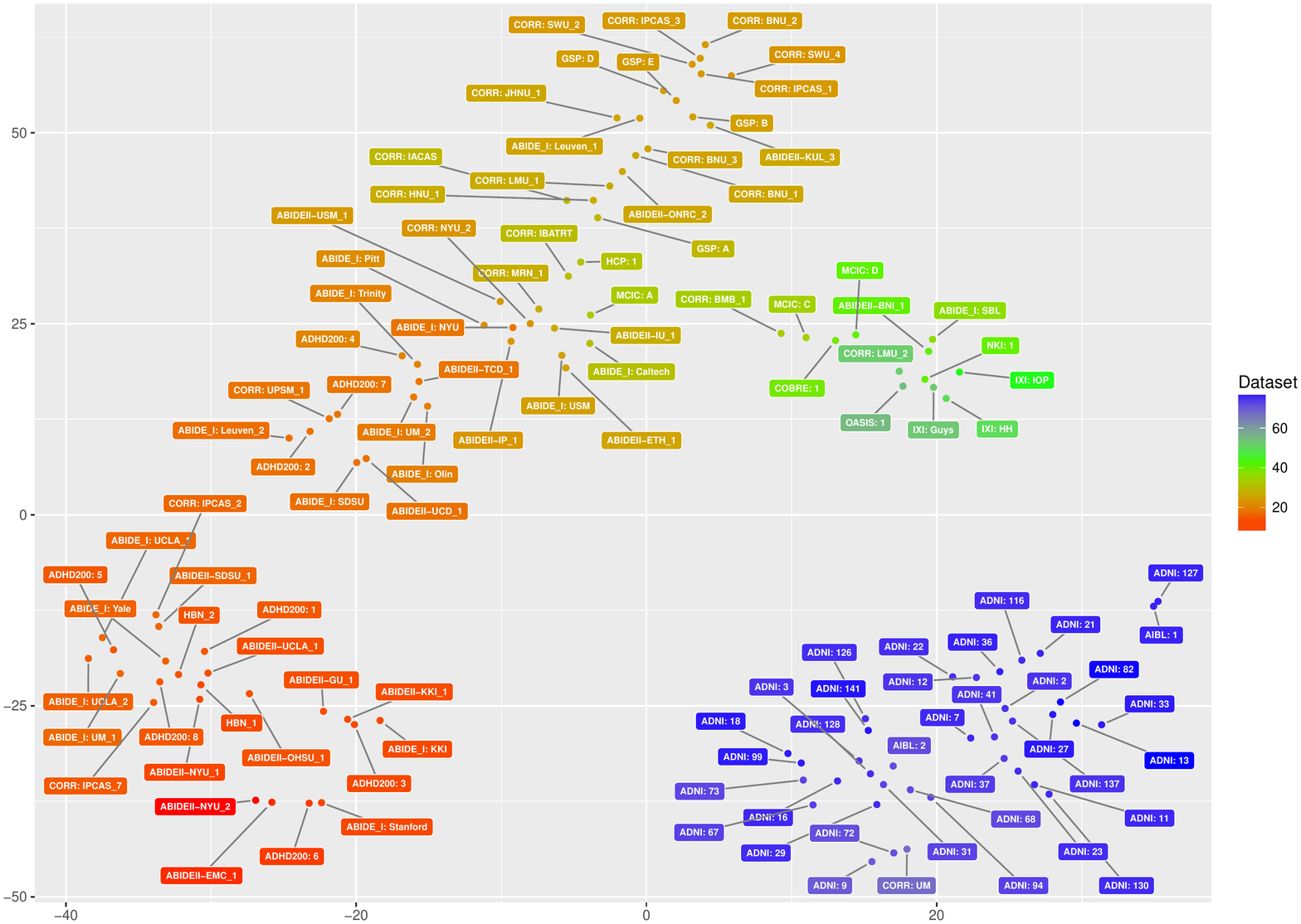} 
\caption{Embedding of neuroimaging sites. Color encodes mean age of site. Name of site and dataset displayed next to data point.  
\label{fig:EmbeddingAge}
}
\end{center}
\end{figure*}

It is striking to see that some sites are more similar to sites from other datasets than to sites from the same dataset. 
We observe four clusters. 
Cluster I contains all sites from ADNI and AIBL, representing old subjects. 
Cluster II consists of sites from IXI, NKI, COBRE, and OASIS, which include subjects from a very wide age range. 
Cluster III has younger subjects mainly in their twenties, including GSP and HCP, together with many sites from ABIDE and CORR. 
Cluster IV mainly contains children and adolescents, e.g., HBN and sites from ABIDE. 
In Fig.~\ref{fig:EmbeddingAge}, we show the same embedding as in Fig.~\ref{fig:Embedding} but with the label color according to the age. 
It is natural to see that the major variations are due to age, due to its predominant impact on brain morphology~\cite{potvin2017normative,Wachinger2015brainprint}.
Within those clusters age is relatively homogeneous so that other factors like field strength and manufacturer can play a role. 
All in all, we believe that such an embedding of the majority of neuroimaging datasets is of great value to clarify the relationship between different datasets. 
In addition, it could be used to guide the combination of data from sites, while limiting the introduction of unwanted variation. 

\subsection{Incorporate Bias in Training Set Selection\label{sec:autism}}
\begin{figure*}[t]
\begin{center}
	\includegraphics[width=0.48\textwidth]{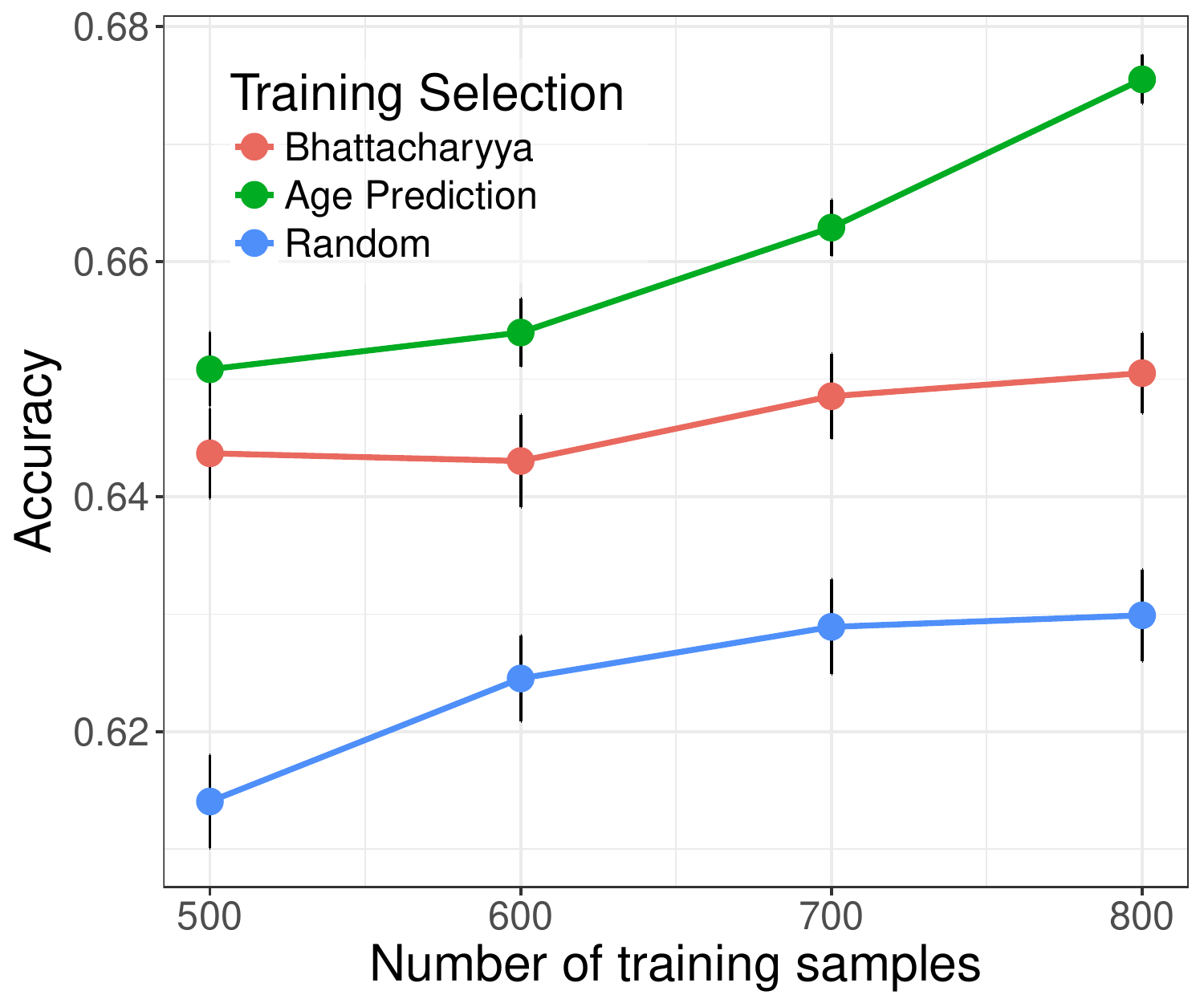} 
	\includegraphics[width=0.48\textwidth]{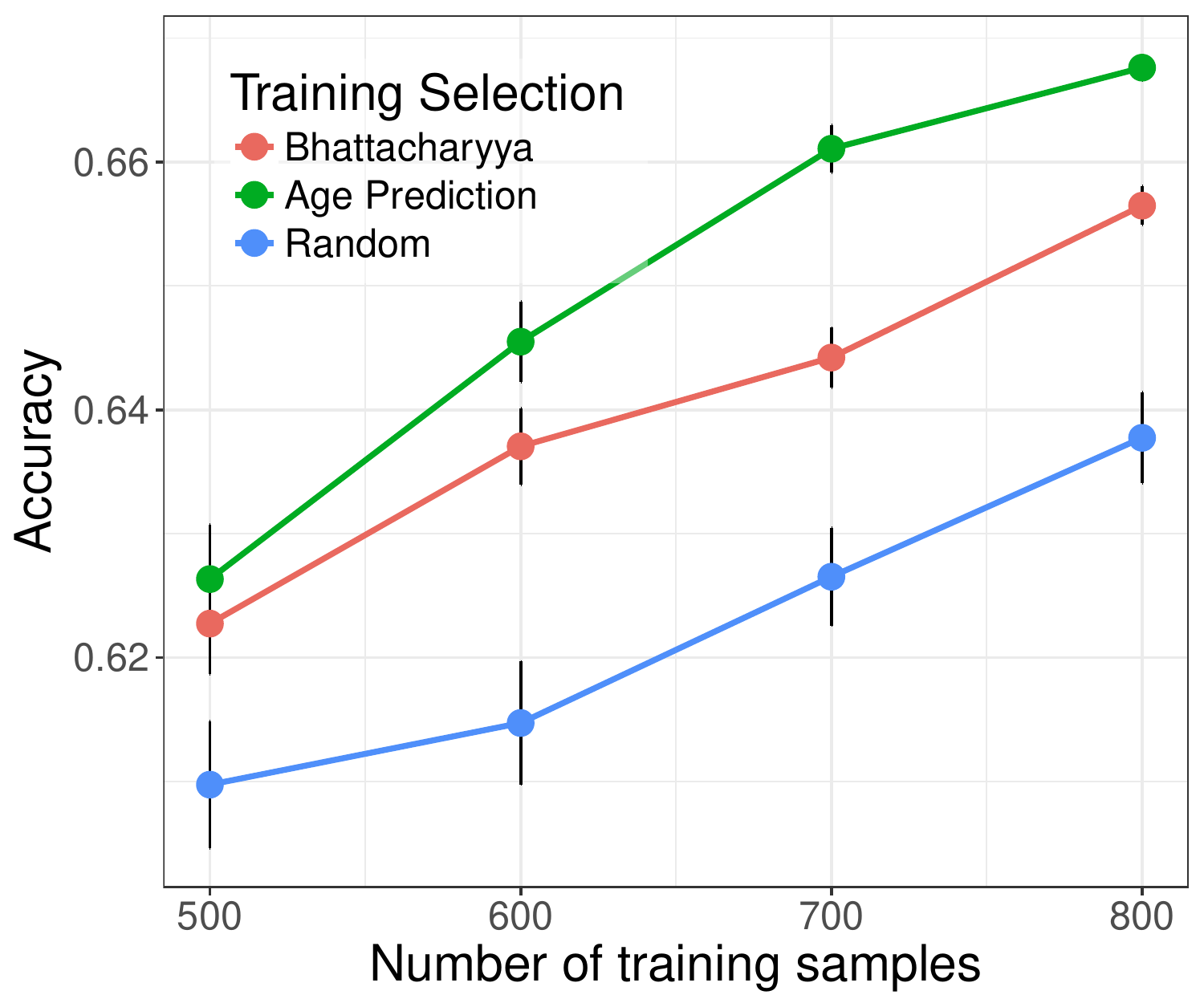} 
\caption{Autism prediction results for site UM from ABIDE I (left) and site KKI from ABIDE II (right) by training set selection with Bhattacharyya distance, age prediction, and random sampling. Curves show the average score over 200 repetitions, error bars show the standard deviation.
\label{fig:Autism}
}
\end{center}
\end{figure*}

We demonstrate the benefits of the compatibility metric for the classification of autism, where we only operate with the ABIDE I + II datasets because the other datasets do not contain autistic subjects. 
To this end, we select one site for testing and we compose the training set based on the metric $m$. 
The rationale is that sites that are close to the target site will be better suited for training a classifier than sites that are very distant. 
In details, we sample the training set from the source dataset that consists of all sites, except for the testing site. 
We encourage the selection of samples from sites that are near by setting the probability of the sample being selected proportional to $\exp(- m(A,B))$, the negative exponential of the site metric. 
As baseline, we use a uniform distribution, which corresponds to random sampling. 
Fig.~\ref{fig:Autism} illustrates the autism classification accuracy for the two largest sites in ABIDE I and ABIDE II, respectively. 
We observe that selecting the training set with either of the two metrics outperforms the random selection, and further that the computation of the distance with age prediction yields the best results. 


Noteworthy, the selection algorithm is  driven by image measurements. 
This makes it on the one hand very versatile, as it can be easily applied to image archives with T1-weighted MRI scans.
On the other hand, by directly operating on the output, this models all of the previously discussed biases. 


%

\section{Conclusion}
On a large collection of datasets with 12,207 individuals, we have illustrated that dataset bias has a strong influence on neuroimaging measures. 
We have quantified dataset compatibility with metrics based on the age prediction error and the Bhattacharyya distance. 
Computation of the metric between all pairs of neuroimaging sites enabled the creation of an embedding, which illustrated that sites across datasets can be more similar than sites within datasets. 
Finally, we demonstrated the advantages of incorporating dataset bias for training set selection in autism prediction, where age prediction outperformed Bhattacharyya distance. 
Overall, we believe that the growing amount of neuroimaging data necessitates to incorporate data-driven methods for quantifying dataset bias in future analyses. 

\noindent
\textbf{Acknowledgement:} This work was supported in part by the  Bavarian State Ministry of Education, Science and the Arts in the framework of the Centre Digitisation.Bavaria (ZD.B).

{\small
\bibliographystyle{splncs03}
\bibliography{jab_bib}
}

\end{document}